\documentclass[runningheads]{llncs}

\usepackage{graphicx}

\usepackage{amsmath}
\usepackage{amssymb}
\usepackage[utf8]{inputenc} % allow utf-8 input
\usepackage[T1]{fontenc}    % use 8-bit T1 fonts
\usepackage{hyperref}       % hyperlinks
\usepackage{url}            % simple URL typesetting
\usepackage{booktabs}       % professional-quality tables
\usepackage{amsfonts}       % blackboard math symbols
\usepackage{graphicx}
\usepackage{nicefrac}       % compact symbols for 1/2, etc.
\usepackage{microtype}      % microtypography
\usepackage{xcolor}         % colors
\usepackage{algorithm} %format of the algorithm 
\usepackage{algorithmic} %format of the algorithm 

\usepackage{adjustbox}
\usepackage{multirow}
\usepackage{xcolor}
\newcommand{\ql}[1]{\textcolor{black}{#1}}

\begin{document}
\title{Partial Least Square Regression via Three-factor SVD-type Manifold Optimization for EEG Decoding}
\titlerunning{PLSR via Three-factor SVD-type Manifold Optimization}
\author{Wanguang Yin\inst{1} \and
Zhichao Liang\inst{1} \and
Jianguo Zhang\inst{2} \and
Quanying Liu\inst{1}}
\authorrunning{PLSR via Three-factor SVD-type Manifold Optimization}
\institute{$^1$ Shenzhen Key Laboratory of Smart Healthcare Engineering, Department of Biomedical Engineering, Southern University of Science and Technology \\
$^2$ Department of Computer Science, Southern University of Science and Technology, Shenzhen, 518055, China \\
\email{\{yinwg,liuqy\}@sustech.edu.cn}}

\maketitle              % typeset the header of the contribution
\begin{abstract}
Partial least square regression (PLSR) is a widely-used statistical model to reveal the linear relationships of latent factors that comes from the independent variables and dependent variables. However, traditional methods \ql{ to solve PLSR models are usually based on the Euclidean space, and easily getting} stuck into a local minimum. To this end, we propose a new method to solve the partial least square regression, named PLSR via optimization on bi-Grassmann manifold (PLSRbiGr). \ql{Specifically, we first leverage} the three-factor SVD-type decomposition of the cross-covariance matrix defined on the bi-Grassmann manifold, converting the orthogonal constrained optimization problem into an unconstrained optimization problem on bi-Grassmann manifold, and then incorporate the Riemannian preconditioning of matrix scaling to regulate the Riemannian metric in each iteration. \ql{PLSRbiGr is validated} with a variety of experiments for decoding EEG signals at motor imagery (MI) and steady-state visual evoked potential (SSVEP) task. Experimental results demonstrate that PLSRbiGr outperforms competing algorithms in multiple EEG decoding tasks, which will greatly facilitate small sample data learning. %\ql{The code of PLSRbiGr is available at \url{https://github.com/ncclabsustech/Riemannian-manifold-optimization-for-PLSR}.}

\end{abstract}
\section{Introduction}

Extracting the latent factor is an essential procedure for discovering the latent space of high-dimensional data; thereby proposed lots of regression model for latent semantic or variable analysis. Among of them, partial least squares regression (PLSR) is a well-established model for learning the latent variables, by a sequential way to learn the latent sub-spaces while maximally maintaining the correlations between the latent variables of independent variables $ X $ and dependent variables $ Y $. Specifically, it can be described as the projection of two variables $ X $ and $ Y $ onto the lower dimensional sub-spaces $ U $ and $ V $. This application can be found in a various of fields, including chemometrics~\cite{brereton2018partial,hasegawa2000rational}, chemical process control~\cite{dong2018regression,zheng2018semisupervised}, and neuroscience~\cite{chu2020decoding,hoagey2019joint}.

As a result, there emerges a lot of partial least square regression models, while most of them are based on the \textit{Euclidean space} to solve the latent factors and identify the latent factors column by column in each iteration. A drawback of this approach is that it easily \textit{converges to a spurious local minimum} and hardly obtain a globally optimal solution. Given an extension of PLSR to the N-way tensors with rank-one decomposition, which can provide an improvement on the intuitive interpretation of the model~\cite{bro1996multiway}. However, N-way PLSR suffers from high computational complexity and slow convergence in dealing with complex data.

To address problems above, we propose a three-factor SVD-type decomposition of PLSR via optimization on bi-Grassmann manifold (PLSRbiGr). Hence, its corresponding sub-spaces $ U $ and $ V $ associated with independent variable $ X $ and dependent variable $ Y $ can be solved by using a nonlinear manifold optimization method. \ql{Moreover, we leverage} the Riemannian preconditioning of matrix scaling $ S^TS $ to regulate the Riemannian metric in each iteration and self-adapt to the changing of subspace~\cite{kasai2016low}. To validate the performance of PLSRbiGr, we conduct two EEG classification tasks on the motor imagery (MI) and steady-state visual evoked potential (SSVEP) datasets. The results demonstrate PLSRbiGr is superior to the inspired modification of PLSR (SIMPLSR) \cite{de1993simpls}, SIMPLSR with the generalized Grassmann manifold (PLSRGGr), SIMPLSR with product manifold (PLSRGStO), as well as the sparse SIMPLSR via optimization on generalized Stiefel manifold (SPLSRGSt) \cite{chen2018solving}. Our main contributions can be summarized as follow:

\begin{itemize}
  \item \ql{We propose a novel method for solving partial least square regression (PLSR), named PLSR via optimization on bi-Grassmann manifold (PLSRbiGr), which decomposes the cross-covariance matrix ($ X^TY $) to an interpretable subspace ($ U $ and $ V $) and simultaneously learns the latent factors via optimization on bi-Grassmann manifold (\textbf{Sec}~\ref{section_PLSR_Riemannian}).}
  \item \ql{We present Riemannian metric that equipped with Riemannian preconditioning, to self-adapt to the changing of subspace. Fortunately, Riemannian preconditioning can largely improves the algorithmic performance (\textit{i.e.} convergence speed and classification accuracy) (\textbf{Sec}~\ref{results:Preconditioning}). }
  \item \ql{The results of EEG decoding demonstrate that PLSRbiGr outperforms conventional Euclidean-based methods and Riemannian-based methods for small sample data learning (\textbf{Sec}~\ref{results:MI}).}
\end{itemize}

\section{Method}

\subsection{Review of Partial Least Squares Regression}
\begin{figure}[t]
      \centering
      \includegraphics[width=0.85\linewidth]{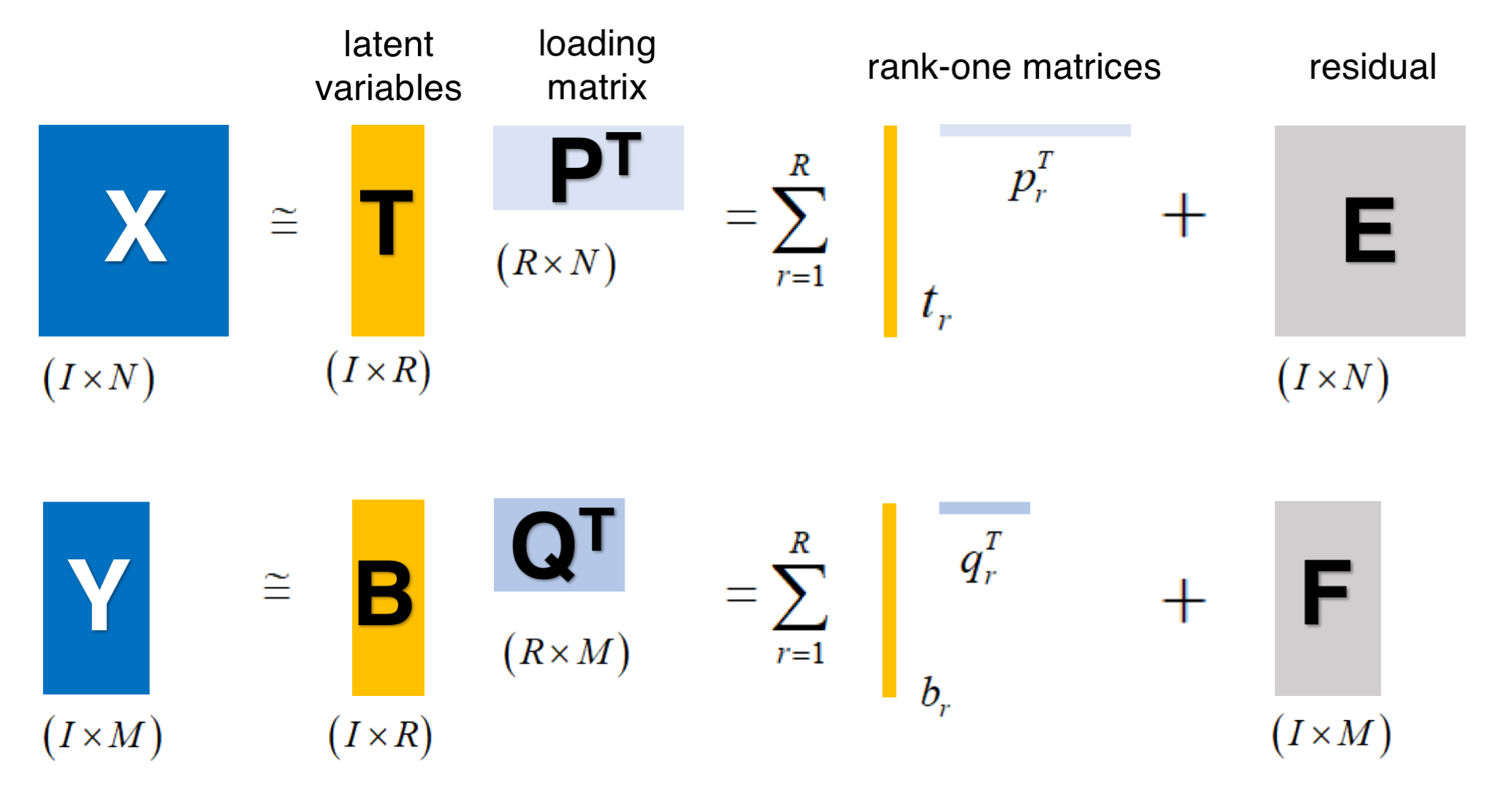}
      \caption{The PLSR decomposes the independent variables (EEG) $ X\in\mathbb{R}^{I\times N} $ and dependent variables (sample labels) $ Y\in\mathbb{R}^{I\times M} $ as a sum of rank-one matrices.}
      \label{fig1}
\end{figure}

Partial least square regression (PLSR) is a wide class of models for learning the linear relationship between independent variables $ X\in\mathbb{R}^{I\times N} $ and dependent variables $ Y\in\mathbb{R}^{I\times M} $ by means of latent variables. We present the schematic illustration of partial least square regression (PLSR) in \textbf{Fig.~\ref{fig1}}. 

To predict dependent variables $ Y $ from independent variables $ X $, PLSR finds a set of latent variables (also called the latent vectors, score vectors, or components) by projecting both $ X $ and $ Y $ onto a lower dimensional subspace while at the same time maximizing the pairwise covariance between the latent variables $ T $ and $ B $, which are presented in Eq.~\eqref{R1} \& Eq.~\eqref{R2}. 

\begin{equation}\label{R1} \small 
X=TP^T + E=\sum_{r=1}^Rt_rp_r^T + E
\end{equation}

\begin{equation}\label{R2} \small
Y=BQ^T + F=\sum_{r=1}^Rb_rq_r^T + F
\end{equation}
where $ T=\left [ t_1,t_2,\ldots,t_R \right ]\in\mathbb{R}^{I\times R} $ is a matrix of $ R $ extracted latent variables from $ X $, and $ B=\left [ b_1,b_2,\ldots,b_R \right ]\in\mathbb{R}^{I\times R} $ are latent variables from $ Y $, their columns have the maximum correlations between each other. In addition, $ P\in\mathbb{R}^{N\times R} $ and $ Q\in\mathbb{R}^{M\times R} $ are the loading matrices, and $ E\in\mathbb{R}^{I\times N} $ and $ F\in\mathbb{R}^{I\times M} $ are the residuals with respect to $ X $ and $ Y $. 

However, \ql{most of the current methods for solving PLSR are based on the Euclidean space by performing the sum of a minimum number of rank-one decomposition to jointly approximate the independent variables $ X $ and dependent variables $ Y $. No existing method solves PLSR with bi-Grassmann manifold optimization. To this end, we propose a novel method for solving PLSR via optimization on the bi-Grassmann manifold.}

\subsubsection{Regression model}
\label{results:taskclassification}
To predict the dependent variable \ql{given a new sample $X_{new}$}, its regression model is given by:

\begin{equation}\label{cost32}
Y^{new}=X_{new}C=X_{new}W\left ( P^TW \right )^{-1}DQ^T
\end{equation}
where $ Y^{new} $ is the predicted output, and $ C=U\left ( P^TU \right )^{-1}DQ^T $ is the regression coefficient calculated from the training data. Following to \cite{chen2018solving}, we can obtain the learned sub-spaces $ U $ \& $ V $, and its corresponding latent factors $ T=XU $ \& $ B=YV $ by solving the SVD-type model via optimization on bi-Grassmann manifold.

\subsection {PLSRbiGr: PLSR via Optimization on Bi-Grassmann Manifold}
\label{section_PLSR_Riemannian}

\begin{figure}[h]
      \centering
      \includegraphics[width=0.85\linewidth]{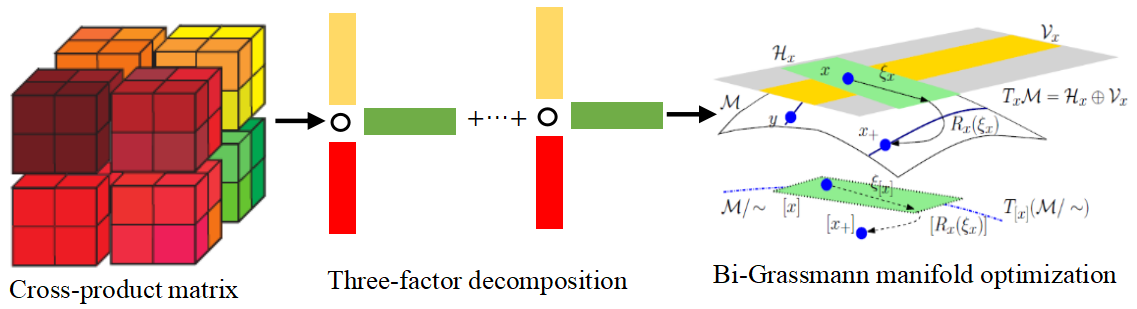}
      \caption{SVD-type decomposition via optimization on bi-Grassmann manifold.}
      \label{Framework}
\end{figure}

In the sequel, we present the SVD-type decomposition via optimization on bi-Grassmann manifold. As shown in \textbf{Fig.~\ref{Framework}}, we treat the cross-product matrix (or tensor) of independent variables and dependent variables as the input data, and then perform the three factor decomposition via optimization on bi-Grassmann manifold.

\subsubsection{Three-factor SVD-type decomposition}
The SVD-type decomposition of cross-product matrix (\textit{i.e.} $ Z $) can be formulated as the following tensor-matrix product:

\begin{equation}\label{cost_objective} \small
f\left ( U,V,S \right ) = \mathop{\mathrm{argmin}}_{S,U,V}\frac{1}{2}\left \| S\times_1U\times_2V - Z\left ( k \right ) \right \|_F^2
\end{equation}
where $ k $ denotes the count of iterations. To apply the Riemannian version of conjugate gradient descent or trust-region method, it needs to obtain the partial derivatives of $f$ with respect to $ U $, $ V $ and $ S $, that is given by

\begin{equation}\label{cost_U} \small
\begin{aligned}
\mathrm{grad}f\left ( U,V,S \right )&=\frac{\partial f\left ( U,V,S \right )}{\partial U}\\
&=\left ( USV^T-Z\left ( k \right ) \right )VS^T\\
&=Z\left ( k+1 \right )VS^T\\
\end{aligned}
\end{equation}

\begin{equation}\label{cost_V} \small
\begin{aligned}
\mathrm{grad}f\left ( U,V,S \right )&=\frac{\partial f\left ( U,V,S \right )}{\partial V}\\
&=\left ( USV^T-Z\left ( k \right ) \right )^TUS^T\\
&=Z\left ( k+1 \right )^TUS^T\\
\end{aligned}
\end{equation}

\begin{equation}\label{cost_S}  \small
\begin{aligned}
\mathrm{grad}f\left ( U,V,S \right )&=\frac{\partial f\left ( U,V,S \right )}{\partial S}\\
&=\left ( S\times_1U\times_2V-Z\left ( k \right ) \right )\times_1U^T\times_2V^T\\
&=U^T\left ( USV^T-Z\left ( k \right ) \right )V\\
&=U^TZ\left ( k+1 \right )V\\
\end{aligned}
\end{equation}
Then, we project the Euclidean gradient (\textit{i.e.} Eq.(\ref{cost_U}), Eq.(\ref{cost_V}), and Eq.(\ref{cost_S})) onto the Riemannian manifold space.

% \section{}
\label{section_bi-Grassmann}
\subsubsection{SVD-type decomposition via optimization on bi-Grassmann manifold}

In this subsection, we derive the bi-Grassmann manifold optimization for the SVD-type decomposition and $ \mathcal{M} $ is a manifold equipped with Riemannian metric. Recall that Stiefel manifold is the set of matrices whose columns are orthogonal, that is denoted by

\begin{equation}\label{cost_Stiefel} \small
St\left ( N,R \right )=\left\{ U\in\mathbb{R}^{N\times R}|U^TU=\mathrm{I}_R \right\}
\end{equation}
For a Stiefel manifold $ St\left ( N,R \right ) $, its related Grassmann manifold $ Gr\left ( N,R \right ) $ can be formulated as the quotient space of $ St\left ( N,R \right ) $, under the equivalence relation defined by the orthogonal group,

\begin{equation}\label{cost7} \small
Gr\left ( N,R \right )=St\left ( N,R \right ) / \mathcal{O}\left ( R \right )
\end{equation}
here, $ \mathcal{O}\left ( R \right ) $ is the orthogonal group defined by

\begin{equation}\label{cost_Quotient} \small
\mathcal{O}\left ( R \right )=\left\{ W\in\mathbb{R}^{R\times R}|W^TW=WW^T=\mathrm{I}_R \right\}
\end{equation}
Moreover, for the SVD-type decomposition, the optimization sub-spaces can be expressed as the following bi-Grassmann manifold:

\begin{equation}\label{cost_Manifold} \small
\mathcal{M}:=Gr\left ( N,R \right )\times Gr\left ( M,R \right )\times \mathbb{R}^{R\times R}
\end{equation}
that is equipped with following Riemannian metric: 

\begin{equation}\label{cost8} \small
\mathrm{g}_Z\left ( \xi _Z,\eta _Z \right )=tr\left ( SS^T\xi _U^T\eta _U \right ) + tr\left ( S^TS\xi _V^T\eta _V \right ) + tr\left ( \xi _S^T\eta _S \right )
\end{equation}
In practice, the computational space (\textit{i.e.} $ Z\in \mathbb{R}^{N\times M} $) can be first decomposed into orthogonal complementary sub-spaces (normal space \textit{i.e.} $ \mathrm{N}_Z\mathcal{M} $, and tangent space \textit{i.e.} $ \mathrm{T}_Z\mathcal{M} $), and then the tangent space can be further decomposed into the other two orthogonal complementary sub-spaces (horizontal space \textit{i.e.} $ \mathrm{H}_Z\mathcal{M} $, and vertical space \textit{i.e.} $ \mathrm{V}_Z\mathcal{M} $), and eventually we project the Euclidean gradient to the horizontal space defined by the equivalence relation of orthogonal group~\cite{absil2009optimization}.

\subsubsection{Riemannian Gradient}
To obtain the Riemannian gradient, it needs to project the Euclidean gradient onto the Riemannian manifold space~\cite{absil2009optimization}, that is

\begin{equation}\label{cost27} \small
\begin{aligned}
\mathrm{Grad}f\left ( U,V,S \right )&=\Pi _U\mathrm{grad}f\left ( U,V,S \right )\\
&=\left ( \mathrm{grad}f\left ( U \right )-UB_U \right )\left ( SS^T \right )^{-1}\\
\end{aligned}
\end{equation}

\begin{equation}\label{cost28} \small
\begin{aligned}
\mathrm{Grad}f\left ( U,V,S \right )&=\Pi _V\mathrm{grad}f\left ( U,V,S \right )\\
&=\left ( \mathrm{grad}f\left ( V \right )-VB_V \right )\left ( S^TS \right )^{-1}\\
\end{aligned}
\end{equation}
where $ \left ( SS^T \right )^{-1} $ and $ \left ( S^TS \right )^{-1} $ is the scaling factors. $ \Pi _U=\mathrm{P}_U^h\mathrm{P}_U^v $ is the project operator that involves two steps of operation, one is a mapping from ambient space to the tangent space $ \left ( \textit{i.e.,}  \mathrm{P}_U^v  \right ) $, and the other is a mapping from tangent space to the horizontal space $ \left ( \textit{i.e.,}  \mathrm{P}_U^h  \right ) $. Therefore, the computational space of Stiefel manifold $ St\left ( N,M \right ) $ can be decomposed into tangent space and normal space, and the tangent space can be further decomposed into two orthogonal complementary sub-spaces (\textit {i.e.} horizontal space and vertical space)~\cite{kasai2016low}. Once the expression of Riemannian gradient and Riemannian Hessian are obtained, we can conduct the Riemannian manifold optimization by using Manopt toolbox~\cite{absil2009optimization}.

\section {Experiments and Results}
\label{section_Experiments}
To test the performance of our proposed algorithm, we conduct experiments on a lot of EEG signal decoding, whose performance is compared to several well known algorithms, including the statistically inspired modification of PLSR (SIMPLSR), SIMPLSR with the generalized Grassmann manifold (PLSRGGr), sparse SIMPLSR via optimization on generalized Stiefel manifold (SPLSRGSt) ~\cite{chen2018solving,de1993simpls}, and higher order partial least squares regression (HOPLSR)~\cite{zhao2012higher}. %The code of PLSRbiGr is available at \url{https://github.com/ncclabsustech/Riemannian-manifold-optimization-for-PLSR}.

\subsection{EEG Decoding}
\label{results:MI}

In this subsection, we test the efficiency and accuracy of our proposed algorithm (PLSRbiGr) on the public PhysioNet MI dataset~\cite{schalk2004bci2000}. We compare PLSRbiGr with other existing algorithms, including the statistically inspired modification of PLSR (SIMPLSR), SIMPLSR with the generalized Grassmann manifold (PLSRGGr), sparse SIMPLSR via optimization on generalized Stiefel manifold (SPLSRGSt) ~\cite{chen2018solving,de1993simpls}, and higher order partial least squares (HOPLSR)~\cite{zhao2012higher}. To evaluate the performance of decoding algorithms, we use Accuracy (Acc) as the evaluation metric to quantify the performance of comparison algorithms.

\begin{figure}[htb]
      \centering
      \includegraphics[width=1.00\linewidth]{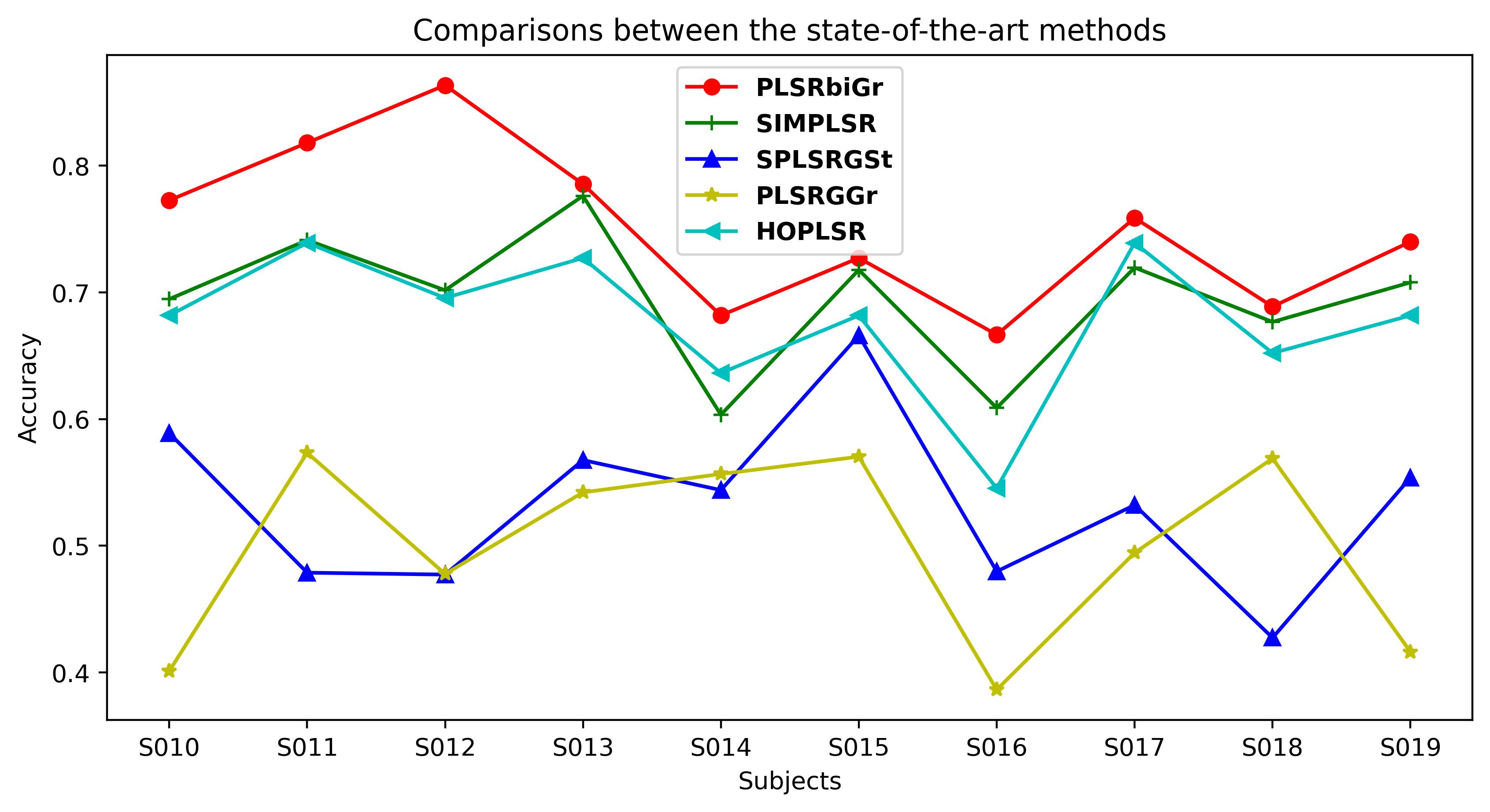}
      \caption{Accuracy of 2-class MI classification task on PhysioNet MI dataset}
      \label{fig2}
\end{figure}

In training, we set the 4-fold-cross-validation to obtain the averaged classification accuracy in testing samples. As shown in \textbf{Fig.~\ref{fig2}}, PLSRbiGr generally achieves the best performance in comparison to the existing methods. The used PhysioNet EEG MI dataset consists of 2-class MI tasks (\textit{i.e.} runs 3, 4, 7, 8, 11, and 12, with imagine movements of left fist or right fist) ~\cite{schalk2004bci2000}, which is recorded from 109 subjects with 64-channel EEG signals (sampling rate equals to 160 Hz) during MI tasks. We randomly select 10 subjects from PhysioNet MI dataset in our experiments. The EEG signals are filtered with a band-pass filter (cutoff frequencies at $ 7\sim35 $ Hz) and a spatial filter (\textit{i.e.} Xdawn with 16 filters), therefore the resulting data is represented by $ trials \times channel \times time $.

\begin{figure}[htb]
      \centering
      \includegraphics[width=1.00\linewidth]{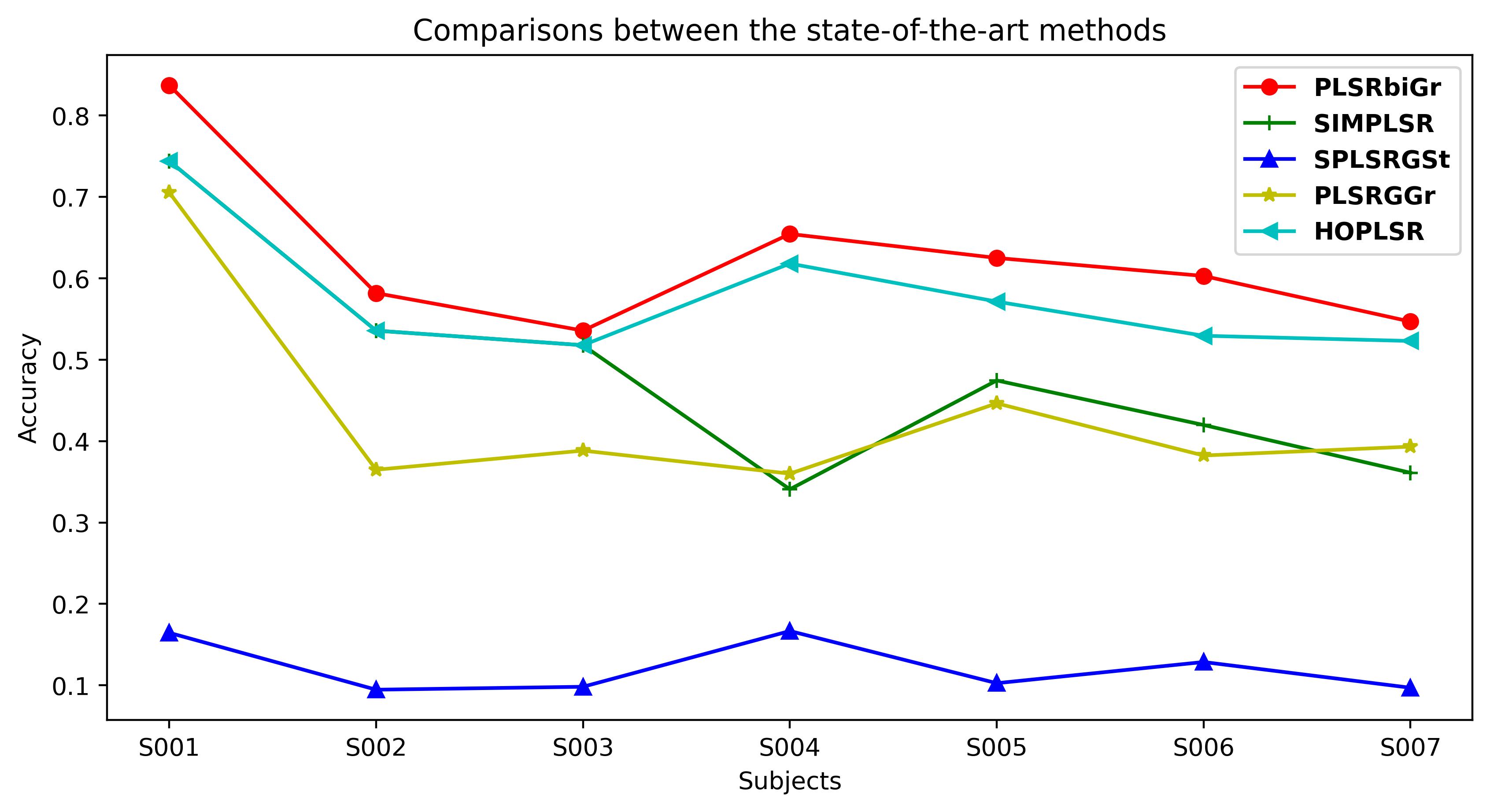}
      \caption{Accuracy of 4-class SSVEP classification task on Macau SSVEP dataset}
      \label{fig3}
\end{figure}

The used Macau SSVEP dataset contains 128-channel EEG recordings from 7 subjects sampled at 1000 Hz, which was recorded at University of Macau with their ethical approval. There are four types visual stimulus with flashing frequency at 10Hz, 12Hz, 15Hz, and 20Hz. To increase the signal-noise ratio (SNR) and reduce the dimension of raw EEG signals, the EEG signals are filtered with a band-pass filter (cutoff frequencies at $ 7\sim35 $ Hz) and a spatial filter (\textit{i.e.} Xdawn with 16 filters). The epochs of 1-second EEG signals before every time point of the SSVEP data were extracted and down-sampled to 200 Hz, therefore the resulting data is represented by $ trial \times channel \times time $. \textbf{Fig.~\ref{fig3}} presents the classification accuracy of all comparison algorithms on the SSVEP dataset. 

\subsection{Effects of Riemannian Preconditioning}
\label{results:Preconditioning}

Furthermore, we test the effects of Riemannian preconditioning. In PLSRbiGr, the scaling factors $ S^TS $ and $ SS^T $ of Riemannian preconditioning provides an effective strategy to accelerate the convergence speed. As shown in Table~\ref{tab:preconditioning}, the classification accuracy and running time of PLSRbiGr equipped with Riemannian preconditioning are closely better than results of other methods, such as PLSRGGr, PLSRGStO, and SPLSRGSt that have not taken into account of Riemannian preconditioning.

\begin{table}[ht]
\centering
	\caption{The effects of Riemannian preconditioning on the classification accuracy and running time. The test experiments were conducted on Macau SSVEP dataset by using PLSRbiGr.}~\label{tab:preconditioning}
		\centering \small
	\setlength{\tabcolsep}{1.2pt}{
\begin{adjustbox}{width=1.00\columnwidth}
\begin{tabular}{c|cc|cc}
\hline
\multirow{2}{*}{ID} & \multicolumn{2}{c|}{Accuracy}    & \multicolumn{2}{c}{Running time (s)} \\ \cline{2-5} 
         & preconditioned        & non-preconditioned  &preconditioned    & non-preconditioned  \\ \hline
 S001    & \bf{0.8487±0.0148}    & 0.3063±0.0394       & 0.0571           & \bf{0.0570}        \\
 S002    & \bf{0.7036±0.0123}    & 0.1211±0.0241       & 0.0624           & \bf{0.0544}        \\
 S003    & \bf{0.5903±0.0183}    & 0.1854±0.0278       & 0.0576           & \bf{0.0551}        \\
 S004    & \bf{0.7360±0.0231}    & 0.1056±0.0258       & 0.0960           & \bf{0.0591}        \\
 S005    & \bf{0.8026±0.0182}    & 0.3187±0.0483       & 0.0692           & \bf{0.0560}        \\
 S006    & \bf{0.7987±0.0075}    & 0.1522±0.0285       & 0.0578           & \bf{0.0574}        \\
 S007    & \bf{0.8375±0.0124}    & 0.2135±0.0375       & 0.0593           & \bf{0.0557}        \\ \hline
 mean    & \bf{0.7596±0.0152}    & 0.2004±0.0330       & 0.0656           & \bf{0.0563}        \\
 \hline
\end{tabular}
\end{adjustbox} }
\end{table}
   
\section{Discussion and Conclusion}
\label{section_conclusion}

In this paper, we propose a novel method, named partial least square regression via optimization on bi-Grassmann manifold (PLSRbiGr) for EEG signal decoding. It features to find the objective solution via optimization on bi-Grassmann manifold. Specifically, to relax the orthogonality constraints of objective function, PLSRbiGr converts the constrained optimization problem in Euclidean space to an optimization problem defined on bi-Grassmann manifold, thereby its corresponding subspaces (\textit{i.e.} $ U $ and $ V $) can be learned by using Riemannian manifold optimization instead of the traditional methods that by deflating the residuals in each iteration. In practice, PLSRbiGr can also be used for image classification, and many other prediction tasks, which has no such limitations. More importantly, extensive experiments on MI and SSVEP datasets suggest that PLSRbiGr robustly outperforms other methods optimized in Euclidean space and has a fast convergence than other algorithms that solved in Riemannian manifold space (Table \ref{tab:preconditioning}), and the scaling factor of Riemannian preconditioning provides a good generalization ability between subjects and robust to variances (Table \ref{tab:preconditioning}).

A limitation of our method is that the column of cross-product matrix equals to the class of training samples, thereby the low-rank nature of PLSRbiGr is a certain value. To address such problem, several different directions can be carried out in our feature work. For example, we only consider the case of cross-product matrix, when cross covariance is a multi-way array (tensors), it needs to further consider the product manifold optimization. For example, the higher order partial least squares (HOPLSR) can also be defined on the Riemannian manifold space~\cite{zhao2012higher}, and directly optimized over the entire product of $ N $ manifolds, thereby the rank of cross-product tensor can be automatically inferred from the optimization procedures. Another issue that needs to be further investigated is how to scale the computation of Riemannian preconditioning to the higher order partial least squares (HOPLSR).

%\section*{Acknowledgements}
%This work was funded in part by the National Key Research and Development Program of China (2021YFF1200800), National Natural Science Foundation of China (62001205), Guangdong Natural Science Foundation Joint Fund (2019A1515111038), Shenzhen Science and Technology Innovation Committee (20200925155957004, KCXFZ2020122117340001, SGDX2020110309280100), Shenzhen Key Laboratory of Smart Healthcare Engineering (ZDSYS20200811144003009).

\bibliographystyle{abbrv}
\bibliography{ref.bib}

\begin{thebibliography}{10}

\bibitem{absil2009optimization}
P.-A. Absil, R.~Mahony, and R.~Sepulchre.
\newblock {\em Optimization algorithms on matrix manifolds}.
\newblock Princeton University Press, 2009.

\bibitem{brereton2018partial}
R.~G. Brereton and G.~R. Lloyd.
\newblock Partial least squares discriminant analysis for chemometrics and
  metabolomics: H ow scores, loadings, and weights differ according to two
  common algorithms.
\newblock {\em Journal of Chemometrics}, 32(4):e3028, 2018.

\bibitem{bro1996multiway}
R.~Bro.
\newblock Multiway calibration. multilinear pls.
\newblock {\em Journal of chemometrics}, 10(1):47--61, 1996.

\bibitem{chen2018solving}
H.~Chen, Y.~Sun, J.~Gao, Y.~Hu, and B.~Yin.
\newblock Solving partial least squares regression via manifold optimization
  approaches.
\newblock {\em IEEE transactions on neural networks and learning systems},
  30(2):588--600, 2018.

\bibitem{chu2020decoding}
Y.~Chu, X.~Zhao, Y.~Zou, W.~Xu, G.~Song, J.~Han, and Y.~Zhao.
\newblock Decoding multiclass motor imagery eeg from the same upper limb by
  combining riemannian geometry features and partial least squares regression.
\newblock {\em Journal of Neural Engineering}, 17(4):046029, 2020.

\bibitem{de1993simpls}
S.~De~Jong.
\newblock Simpls: an alternative approach to partial least squares regression.
\newblock {\em Chemometrics and intelligent laboratory systems},
  18(3):251--263, 1993.

\bibitem{dong2018regression}
Y.~Dong and S.~J. Qin.
\newblock Regression on dynamic pls structures for supervised learning of
  dynamic data.
\newblock {\em Journal of Process Control}, 68:64--72, 2018.

\bibitem{hasegawa2000rational}
K.~Hasegawa, M.~Arakawa, and K.~Funatsu.
\newblock Rational choice of bioactive conformations through use of
  conformation analysis and 3-way partial least squares modeling.
\newblock {\em Chemometrics and Intelligent Laboratory Systems},
  50(2):253--261, 2000.

\bibitem{hoagey2019joint}
D.~A. Hoagey, J.~R. Rieck, K.~M. Rodrigue, and K.~M. Kennedy.
\newblock Joint contributions of cortical morphometry and white matter
  microstructure in healthy brain aging: a partial least squares correlation
  analysis.
\newblock {\em Human brain mapping}, 40(18):5315--5329, 2019.

\bibitem{kasai2016low}
H.~Kasai and B.~Mishra.
\newblock Low-rank tensor completion: a riemannian manifold preconditioning
  approach.
\newblock In {\em International Conference on Machine Learning}, pages
  1012--1021. PMLR, 2016.

\bibitem{schalk2004bci2000}
G.~Schalk, D.~J. McFarland, T.~Hinterberger, N.~Birbaumer, and J.~R. Wolpaw.
\newblock Bci2000: a general-purpose brain-computer interface (bci) system.
\newblock {\em IEEE Transactions on biomedical engineering}, 51(6):1034--1043,
  2004.

\bibitem{zhao2012higher}
Q.~Zhao, C.~F. Caiafa, D.~P. Mandic, Z.~C. Chao, Y.~Nagasaka, N.~Fujii,
  L.~Zhang, and A.~Cichocki.
\newblock Higher order partial least squares (hopls): a generalized multilinear
  regression method.
\newblock {\em IEEE transactions on pattern analysis and machine intelligence},
  35(7):1660--1673, 2012.

\bibitem{zheng2018semisupervised}
J.~Zheng and Z.~Song.
\newblock Semisupervised learning for probabilistic partial least squares
  regression model and soft sensor application.
\newblock {\em Journal of process control}, 64:123--131, 2018.

\end{thebibliography}
%\end{thebibliography}
\end{document}